\pdfoutput=1

\documentclass[letterpaper]{article} 
\usepackage{aaai25}  
\usepackage{times}  
\usepackage{helvet}  
\usepackage{courier}  
\usepackage[hyphens]{url}  
\usepackage{graphicx} 
\usepackage{natbib}  
\usepackage{caption} 
\usepackage{algorithm}
\usepackage{algorithmic}

\usepackage{newfloat}
\usepackage{listings}
\DeclareCaptionStyle{ruled}{labelfont=normalfont,labelsep=colon,strut=off} 
\floatstyle{ruled}
\newfloat{listing}{tb}{lst}{}
\floatname{listing}{Listing}
\usepackage{amssymb}
\usepackage{subcaption}
\usepackage{amsmath}
\usepackage{multirow}
\usepackage{textgreek}
\usepackage{hyperref}
\usepackage{textcomp}

\nocopyright

\title{QCS: Feature Refining from Quadruplet Cross Similarity for Facial Expression Recognition}
\author {
    Chengpeng Wang\textsuperscript{\rm 1},
    Li Chen\textsuperscript{\rm 3},
    Lili Wang\textsuperscript{\rm 3},
    Zhaofan Li\textsuperscript{\rm 4},
    Xuebin Lv\textsuperscript{\rm 2}\thanks{Corresponding Author. Email: lvxb@scu.edu.cn}
}
\affiliations {
    \textsuperscript{\rm 1}Wisesoft Inc., Chengdu, China\\
    \textsuperscript{\rm 2}School of Computer Science Sichuan University, Chengdu, China\\
    \textsuperscript{\rm 3}Chinese PLA General Hospital, Beijing, China\\
    \textsuperscript{\rm 4}Medical School of Chinese PLA, Beijing, China\\
}

\usepackage{bibentry}

\begin{document}

\maketitle

\begin{abstract}
Facial expression recognition faces challenges where labeled significant features in datasets are mixed with unlabeled redundant ones. In this paper, we introduce Cross Similarity Attention (CSA) to mine richer intrinsic information from image pairs, overcoming a limitation when the Scaled Dot-Product Attention of ViT is directly applied to calculate the similarity between two different images. Based on CSA, we simultaneously minimize intra-class differences and maximize inter-class differences at the fine-grained feature level through interactions among multiple branches. Contrastive residual distillation is utilized to transfer the information learned in the cross module back to the base network. We ingeniously design a four-branch centrally symmetric network, named Quadruplet Cross Similarity (QCS), which alleviates gradient conflicts arising from the cross module and achieves balanced and stable training. It can adaptively extract discriminative features while isolating redundant ones. The cross-attention modules exist during training, and only one base branch is retained during inference, resulting in no increase in inference time. Extensive experiments show that our proposed method  achieves state-of-the-art performance on several FER datasets.
\end{abstract}

%
\begin{links}
    \link{Code}{https://github.com/birdwcp/QCS}
\end{links}

\section{Introduction}
Facial expression is crucial for machines to comprehend human emotions. Facial expression recognition (FER) holds vast research potential and application worth in human-computer interaction, emotion analysis, and intelligent healthcare. Over decades, deep learning advancements in computer vision have propelled significant progress in facial expression recognition.

\begin{figure}[t]
	\centering
	\includegraphics[width=8cm]{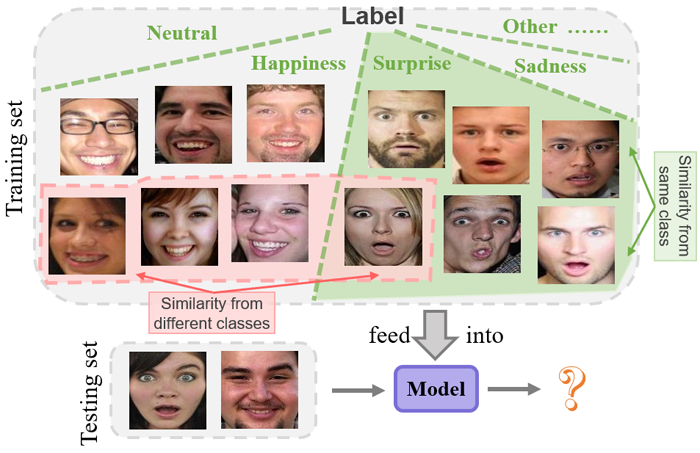}
	\caption{We often assume labeled features exhibit significant and dominant distributions within the dataset; however, unlabeled redundant features may impact this. Feature similarity can be leveraged to distinguish target features from redundant ones, based on whether they originate from the same label group or different label groups.}
	\label{fig:f1}
\end{figure}

However, FER is still a complex task facing many difficulties such as Inter-class similarity and intra-class variances. Faces with varying expressions retain the common features that constitute a human face, yet same expressions can vary widely in pose, identity, illumination, etc. Traditional object classification methods struggle with complex face datasets due to assumptions of labeled feature dominance. Dividing datasets by labels for training can result in the learning of unlabeled redundant features if they're non-randomly distributed, impacting classification accuracy. Fig.~\ref{fig:f1} illustrates one such impact. In reality, interference features are diverse, numerous, and unpredictable, exerting varying degrees of influence.

How to extract discriminative features and how to eliminate the interference of redundant features are the core issues of FER. Using self-attention, some deep learning methods \cite{Kai2020Region,fanglei2021transfer,fanglei2022vision,Fuyan2021Facial} focus on key features, enhancing model discrimination. However, model performance hinges on the high-quality of training data. Some methods \cite{Siyue2019deepmultipath,Weizhang2021Learning,Jiecai2021IdentityFree,Delian2021Feature,Delian2022Adaptive} study separating redundant from expression features using GANs or transformers. They extract expression information via decomposition and reconstruction, but often need redundant feature labels, such as age, gender, race, or rely on identity recognition modules. Additionally, interference decoupling FER \cite{shanli2017reliable} relies on one or several interference feature labels and can only address limited and specific interference factors. However, interference features are often unclear in type and quantity, and these methods that rely on labeled data often fail to achieve good performance when the types, quantities, and quality of labeled data are limited. Furthermore, when the expression feature labels are ambiguous \cite{Jiahui2021DiveintoAmbiguity,Yuhang2021Relative}, it is unrealistic to clearly label all interference features.

Inspired by self-supervised contrastive learning methods\cite{yang2022mcl,Zheng2024Self} that extract feature representations from unlabeled data, we attempt to conduct similarity comparisons between image pairs of different classes in order to isolate these redundant and interfering features, even though they are unlabeled. Additionally, we harness the similarity of same-class image pairs to extract more discriminative common features. Our method differs from those \cite{YuGao2020Channel,Peiqin2020Learning} that utilize channel similarity and only perform comparisons between two images. It is also distinct from approaches\cite{yang2022mcl,Raviteja2019ACompact} that solely contrast high-level features among triplets.

Firstly, we propose Cross Similarity Attention (CSA), specifically tailored for mining fine-grained feature similarity between different images. We have identified a limitation when the Scaled Dot-Product Attention (SDPA) from ViT \cite{Alexey2021ViT} is directly applied to calculate image similarity: the similarity weights derived from the correlation between Q (query) and K (key) directly act on the feature vectors of the paired image, rather than the output feature vectors of the branch, resulting in attention that cannot obtain direct feedback (See Fig.~\ref{fig:f3_b} for details). Our CSA focuses solely on the importance of fine-grained similarity for the K (key), and directs attention directly between the input and output feature vectors of one branch. To our knowledge, we are the first to implement global spatial attention in the spatial dimension, analogous to the method of SENet\cite{JieHu2018Squeeze} in the channel dimension. Previous works, such as BAM\cite{Jongchan2018BAM} and CBAM\cite{Sanghyun2018CBAM}, were limited to capturing only local spatial attention. The self-attention in ViT does not output a complete attention map after applying the softmax function; instead, it outputs results for each individual feature point.

By leveraging CSA for the interaction module, we simultaneously minimize intra-class feature differences and maximize inter-class differences, mimicking the effect of Triplet loss \cite{Florian2015FaceNet} at the fine-grained feature level. In this way, our method enhances discriminative features and separates redundant ones. To effectively integrate the refined information from cross modules, we introduce residual connections between the cross classifier and the base classifier. To avoid gradient conflicts arising from multiple interaction modules across the anchor, positive, and negative branches, we introduce an additional image of the same category as the negative to construct a four-branch centrally symmetric closed-loop network. This framework reduces the computation of two interaction modules and enables a stable joint training. During inference, the multi-branch interaction modules are removed, leaving only a single base classifier, without increasing the inference time.

In summary, our main contributions are summarized as follows:
\begin{itemize}
	\item We introduce Cross Similarity Attention (CSA), a global spatial attention mechanism specifically designed to mine fine-grained feature similarities across different images. It addresses the issue of indirect attention between input and output vector of one branch when computing similarity within the cross attention of ViT.
	\item We propose Quadruplet Cross Similarity (QCS), a novelty four-branch, centrally symmetric closed-loop joint training framework based on CSA. It refines features by adaptively mining discriminative features within the same class while simultaneously separating interfering features across different classes.
	\item Compared to methods that introduce additional landmark information, our method achieves state-of-the-art performance on several FER datasets by mining richer intrinsic information.
\end{itemize}

\begin{figure*}[t]  
	\centering  
	\includegraphics[width=1\textwidth]{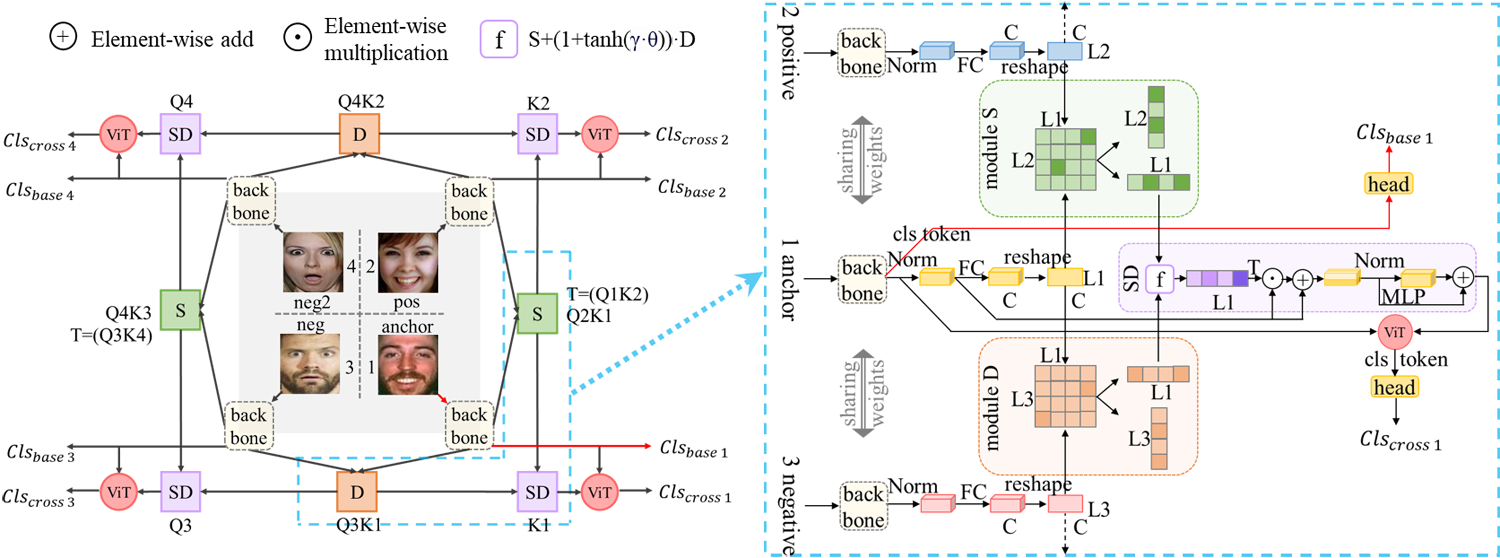}  
	\caption{The framework of Quadruplet Cross Similarity Network. Joint training is performed on 4 classifiers ${{Cls}_{base}}$ based on a weight-shared backbone and 4 classifiers ${{Cls}_{cross}}$ enhanced by cross module, with only the red branch retained during inference. Anchor and pos are in the same category, so are neg and neg2. The matrix ${SD}$ performs attention refinement on matrix ${S}$ and matrix ${D}$ by rows or columns.}  
	\label{fig:framework}  
\end{figure*}

\section{Related Work}
\subsection{Facial Expression Recognition}
With the development of deep learning, FER has become increasingly popular in recent years, as more and more researchers focus on human-computer interaction. RAN \cite{Kai2020Region} proposes a Region Attention Network that adaptively captures the importance of different region features. SCN\cite{Kai2020Suppressing} introduces the Self-Cure Network, which suppresses uncertainty in facial expression datasets by excluding samples with ambiguous annotations. DMUE\cite{Jiahui2021DiveintoAmbiguity} leverages image pair similarity comparisons to mine real label confidence and potential distribution information. RUL\cite{Yuhang2021Relative} computes relative uncertainty through image comparisons, addressing uncertainty arising from blurry images and inaccurate labels.

Transfer\cite{fanglei2021transfer} is the first attempt to apply Vision Transformer (ViT) to adaptively characterize the relations between different facial parts. APViT\cite{fanglei2022vision} proposes two Attentive Pooling modules with Vision Transformer to focus on the most discriminative features and discard the less relevant ones. POSTER\cite{Zheng2023Poster}, POSTER V2\cite{Jiawei2023POSTERV2}, LA-Net\cite{Zhiyu2023LA-Net}, and S2D\cite{Yin2023From} utilize ViT cross-fusion to integrate information from a facial landmark detector MobileFaceNet\cite{MobileFaceNet2021}, while FMAE\cite{ning2024representation} is pretrained on a dataset of 9 million facial images (termed Face9M) for tasks related to AU detection and FER.

\subsection{Attention Mechanisms}
Self-attention primarily captures dependencies within a sequence to focus on important information.  SENet\cite{JieHu2018Squeeze} introduced a plug-and-play channel attention module. BAM\cite{Jongchan2018BAM} and CBAM\cite{Sanghyun2018CBAM} introduced spatial attention modules based on convolutions. However, these convolution-based spatial attention mechanisms can only perform local information interaction. The Non-Local Block\cite{Xiaolong2018Nonlocal} can capture long-range dependencies between any two positions in images. Subsequently, the Transformer\cite{Ashish2017Attention} based on Scaled Dot-Product Attention became the state-of-the-art for natural language processing (NLP) tasks. ViT\cite{Alexey2021ViT} was the first to implement a pure transformer structure in the field of computer vision.

Using distinct features for computing interaction relationships constitutes cross-attention. In Transformer, K and V sequences stem from the source encoder, while Q sequence originates from the target decoder. In Few-shot classification tasks, cross-attention in CEC and CAN\cite{Jinxiang2023Clustered,Ruibing2019Cross} can extract more discriminative features even when the sample size is limited. Fine-grained image classification methods CIN\cite{YuGao2020Channel} and API\cite{Peiqin2020Learning} propose channel cross-attention, which is used to extract channel similarities between image pairs. CrossViT\cite{Chun2021CrossViT} unifies self-attention features across different scales by fusing information from small-patch and large-patch tokens. In multimodal feature fusion\cite{XiWei2020Multi-Modality,Arsha2021Attention,Yiyuan2023MetaTransformer}, cross-attention is used to exchange information between different modalities through early, middle, or late fusion.

\subsection{Contrastive Learning}
Contrastive learning is a type of Self-Supervised Learning method\cite{Ting2020ASimple,Xinlei2021AnEmpirical}. Supervised Contrastive Learning\cite{Prannay2020Supervised} introduces labeled and different data of the same class for contrastive learning, improving the discriminative ability of the model. Triplet loss\cite{Florian2015FaceNet} is a commonly used contrastive learning loss function in face recognition tasks. FEC\cite{Raviteja2019ACompact} learns a compact and continuous expression embedding space by comparing the similarity of three images using Triplet loss.


\section{Methods}
\subsection{Cross Similarity Attention}
Intuitively, comparing the similarities and differences between images can make it easier to judge the category of an image. This similarity comes not only from high-level semantic features but also from relatively low-level fine-grained features. Based on the non-local block\cite{Xiaolong2018Nonlocal}, we propose a spatial attention called Cross Similarity Attention (CSA) to focus on the similarity between different images. This is significantly different from the Scaled Dot-Product Attention in ViT\cite{Alexey2021ViT}, and it does not require concat on spatial dimension of different images like the non-local block.

\begin{figure}[h!]  
	\centering  
	\begin{subfigure}[b]{0.22\textwidth} 
		\includegraphics[width=\textwidth]{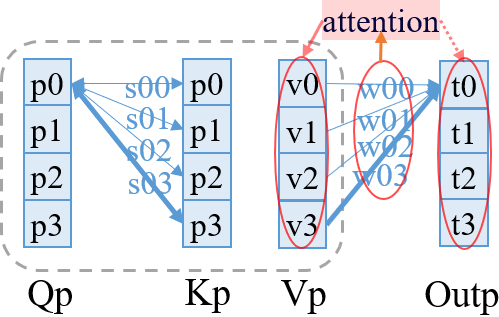}  
		\caption{Self Attention}  
		\label{fig:f3_a}  
	\end{subfigure} 
	\vspace{1mm}  
	\hspace{3mm}
	\begin{subfigure}[b]{0.22\textwidth}  
		\includegraphics[width=\textwidth]{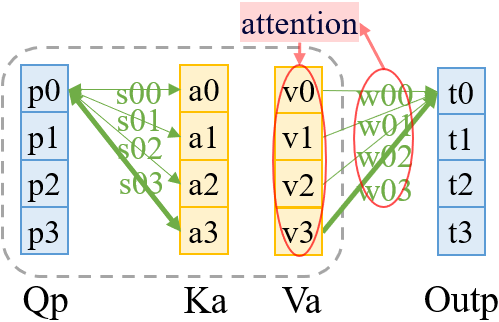}  
		\caption{Cross Attention}  
		\label{fig:f3_b}  
	\end{subfigure}
	\begin{subfigure}[b]{0.22\textwidth}  
		\includegraphics[width=\textwidth]{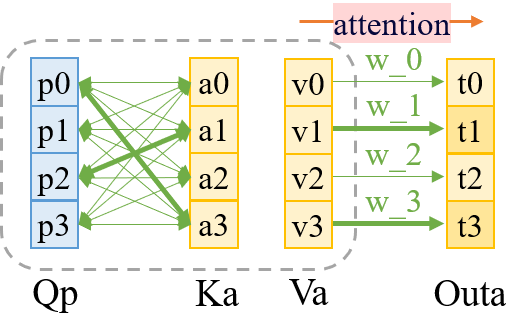}  
		\caption{ \centering Cross Similarity Attention}  
		\label{fig:f3_c}  
	\end{subfigure} 
	\hspace{3mm}
	\begin{subfigure}[b]{0.22\textwidth}  
		\includegraphics[width=\textwidth]{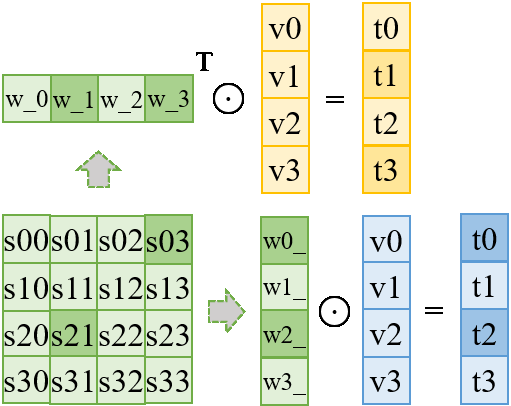}  
		\caption{Similarity Matrix}  
		\label{fig:f3_d}  
	\end{subfigure}  
	\caption{Different types of attention on feature vectors h${\times}$w${\times}$c, where h=w=2, and c denotes channels. Blue and yellow represent features from different images, and each element corresponds to a 1${\times}$1${\times}$c vector. Double-arrowed lines denote the interaction values between corresponding two elements, and thicker lines indicate higher weights. The 4x4 green box in (d) represents the interaction matrix of (c).}
	\label{fig:f3}  
\end{figure}  

Firstly, In standard ViT qkv self-attention (as shown in Fig.~\ref{fig:f3_a}), the query ${Q}$, key ${K}$, and value ${V}$ are all feature vectors derived from the same image. The attention weights acting on the queried vectors are equivalent to acting on the output vectors.

However, if the query ${Q}$ comes from the features of one image p, while the key ${K}$ and values ${V}$ are derived from the features of a different image a, as shown in Fig.~\ref{fig:f3_b}, some changes occur. The output feature points in image ${a}$ are different from the query feature points in image ${p}$ which makes the output feature points unrelated to the calculation of the weights. The weight ${w}$ applies attention to the values ${V}$ and outputs to a feature point in the vector ${Out}$. The attention distribution within the vector ${Out}$ is difficult to assess through the weights.

Therefore we consider only quantifying the importance of the positions in the queried image (key ${K}$), regardless of where the query ${Q}$ comes from. Using these weights, we directly assign weight to all the features of values ${V}$ and output them to the vector ${Out}$ of the same branch, as shown in Fig.~\ref{fig:f3_c} \ref{fig:f3_d}.
\begin{figure}[h]  
	\centering  
	\begin{subfigure}[t]{0.16\textwidth} 
		\includegraphics[width=\textwidth]{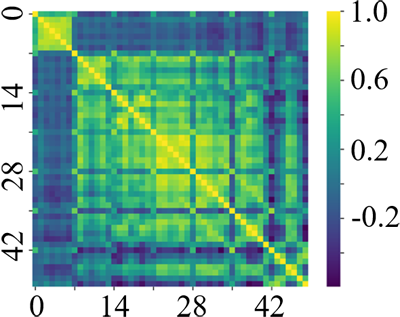}  
		\caption{\centering cosine similarity SA}  
		\label{fig:f4_a}  
	\end{subfigure}    
	\begin{subfigure}[t]{0.16\textwidth}  
		\includegraphics[width=\textwidth]{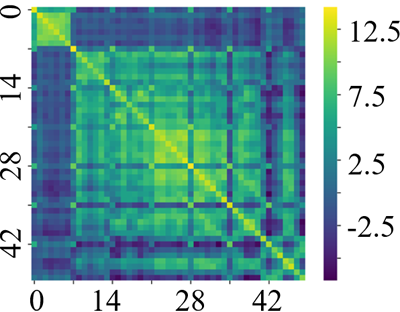}  
		\caption{\centering dot product SA}  
		\label{fig:f4_b}  
	\end{subfigure}  
	\begin{subfigure}[t]{0.16\textwidth}  
		\includegraphics[width=\textwidth]{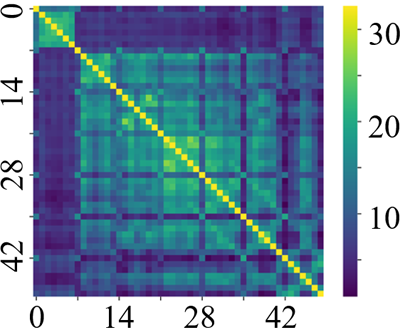}  
		\caption{\centering euclidean distances SA}  
		\label{fig:f4_c}  
	\end{subfigure}  
	\begin{subfigure}[t]{0.16\textwidth}  
		\includegraphics[width=\textwidth]{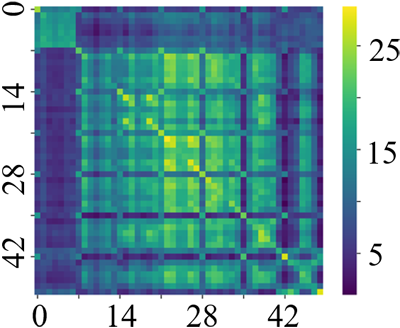}  
		\caption{\centering euclidean distances CA}  
		\label{fig:f4_d}  
	\end{subfigure}  
	\caption{Interaction matrix of 7${\times}$7${\times}$c (h=7, w=7, channel=c) image features in the spatial dimension (h${\times}$w=49).}
	\label{fig:f4}  
\end{figure}  

Secondly, the self-attention in the standard ViT tends to capture the interactions rather than the similarity between vectors. The $D_{h}$ in ${W_{s}\left( {Q,K} \right)=softmax(}{{{Q}{K}^{{T}}}/\sqrt{{D}_{{h}}}}{) }$  is a unified scaling factor. If we emphasize similarity, when calculating self-attention(SA), the vector at a certain spatial position in the same image should have a higher response with itself, as shown in Fig.~\ref{fig:f4_a} \ref{fig:f4_c}. The yellow on the diagonal of the cosine similarity or Euclidean distance similarity matrix is more prominent. However, the dot product (Fig.~\ref{fig:f4_b}) calculation used in ViT weakens the response intensity on the diagonal of the interaction matrix. Meanwhile, the similarity matrix (Fig.~\ref{fig:f4_d}) between different images based on cross-attention(CA) lacks a highlighted diagonal, where ${\text{S}_{\text{c}\text{s}}\left( \text{Q,K} \right)}_{\text{ij}}\neq{\text{S}_{\text{c}\text{s}}\left( \text{Q,K} \right)}_{\text{ji}}$. 

Meanwhile, although the self-attention mechanism of the original ViT module can capture similarities among concatenated feature vectors, it also introduces interference from the similarities among the elements within the feature representation of a single image. The concat operation enlarges the interaction matrix. In contrast, our CSA (Figure~\ref{fig:f3_c}) computes similarities between different feature maps more efficiently.

\begin{small}
	\begin{equation}  
		D_{cs\_pa}\left( Q_{p},K_{a} \right) = \parallel Q_{p} - K_{a} \parallel \in \mathbb{R}^{l \times l} \label{eq:3} 
	\end{equation} 
	\begin{equation} 
		\begin{split}  
			S_{cs\_pa}\left( Q_{p},K_{a} \right) = D_{cs\_pa}\left( Q_{p},K_{a} \right)_{\max} \\
			- D_{cs\_pa}\left( Q_{p},K_{a} \right) \in \mathbb{R}^{l \times l} \label{eq:4}
		\end{split} 
	\end{equation} 
\end{small}	

We calculate the similarity interaction matrix using Euclidean distance as shown in Eq.~\ref{eq:3} \ref{eq:4}, ${D_{cs}\left( {Q,K} \right)_{ij} = \sqrt{\sum_{c=1}^{C} (Q_{ic} - K_{jc})^2} }$, ${Q,K,V \in \mathbb{R}^{l \times c}}$ where ${l = h \times w}$) and separately assess the importance of the queried positions in both images, thereby emphasizing the weights of common features with high similarity across different spatial locations within each image. Inspired by GCNet\cite{Yue2019GCNet}, which observed that attention maps calculated for different query positions are nearly identical, we aggregate the interaction matrix by rows or columns.
\begin{small}
	\begin{equation} 
		S_{cs\_a}\left( Q_{p},K_{a} \right) = \sum_{i = 1}^{l} \left( \operatorname{norm}\left( \left(S_{cs\_pa}\left( Q_{p},K_{a} \right)\right)_{i,:} \right) \right)_{ij} \in \mathbb{R}^{1 \times l} \label{eq:5}
	\end{equation}  
	\begin{equation} 
		A_{cs\_a}\left( Q_{p},K_{a} \right) = \operatorname{softmax}\left( S_{cs\_a}\left( Q_{p},K_{a} \right) \right)^{T} \cdot V_{a} \in \mathbb{R}^{l \times 1} \label{eq:6}
	\end{equation} 
\end{small}	
Eq.~\ref{eq:5} \ref{eq:6} calculate the attention for image ${a}$. Here, ${\operatorname{norm}\left( {S_{cs}\left( {Q,K} \right)}_{i,:} \right)}$ normalizes the similarity matrix by row using the ${L2}$ norm to eliminate absolute differences in similarity calculated from query features at different positions, then sums the values for the same queried spatial position across columns. Finally, a mutually exclusive softmax function is employed to directly compute the weight distribution along the spatial dimension, where ${\sum_{j = 1}^{l}\left( \operatorname{softmax}\left( {S_{cs\_ a}\left( {Q,K} \right)} \right)^{T} \right)_{j}}=1$. We use softmax in the global spatial dimension because it operates on the similarity map of two different images, rather than on the feature map of a single image. Thus, there is no concern that applying Eq.~\ref{eq:6} will disrupt the semantic structure of the image. Eq.~\ref{eq:7} \ref{eq:8} symmetrically calculate the attention for image ${p}$.
\begin{small}
	\begin{equation} 
		S_{cs\_p}\left( Q_{p},K_{a} \right) = \sum_{j = 1}^{l} \left( \operatorname{norm}\left( \left(S_{cs}\left( Q_{p},K_{a} \right)\right)_{:,j} \right) \right)_{ij} \in \mathbb{R}^{l \times 1} \label{eq:7}
	\end{equation} 
	\begin{equation} 
		A_{cs\_p}\left( Q_{p},K_{a} \right) = \operatorname{softmax}\left( S_{cs\_p}\left( Q_{p},K_{a} \right) \right) \cdot V_{p} \in \mathbb{R}^{l \times 1} \label{eq:8} 
	\end{equation}  
\end{small}

\subsection{Quadruplet Cross Similarity Network}

\begin{figure}[t]  
	\centering  
	\includegraphics[width=6cm]{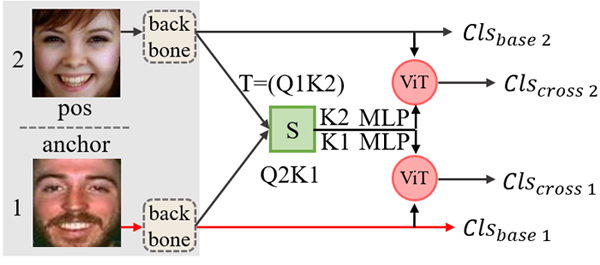}  
	\caption{The framework of Dual Cross Similarity Network.}  
	\label{fig:dcs}  
\end{figure}

By referencing the method of interactive comparison between two images \cite{YuGao2020Channel,Peiqin2020Learning}, we utilize Cross Similarity Attention to enhance fine-grained features that exhibit high similarity between two images belonging to the same category. The resulting classifier, ${{Cls}_{cross}}$ , outperforms the backbone-based ${{Cls}_{base}}$ in discrimination. This approach is termed the Dual Cross Similarity (DCS) Network (Fig.~\ref{fig:dcs}).

\begin{figure}[t]  
	\centering  
	\includegraphics[width=6cm]{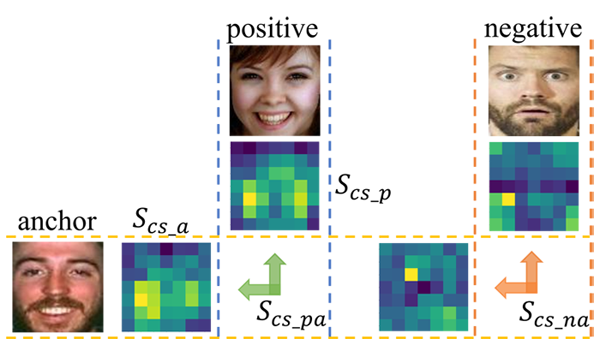}  
	\caption{cross similarity heatmap. yellow represents higher similarity, and blue represents greater distance, i.e., lower similarity. The cross similarity vectors ${{S}_{cs\_ p} (1\times49)}$ and ${{S}_{cs\_ a} (49\times1)}$ are reshaped to  ${7\times7}$ heatmaps.}  
	\label{fig:f3_2_heatmap}  
\end{figure}

However, merely relying on Cross Similarity Attention for feature weighting in same-category image pairs has limitations. Highly similar features may not be discriminative, e.g., flat edge areas in most face images. Unlabeled features like gender, skin tone, etc., may exhibit similarity between images of the same or different categories. Leveraging distribution differences with labeled features via cross similarity attention can mine redundant interfering features, separating them from similar same-category features, refining cleaner features. The cross similarity heatmaps between the anchor image and images of the same or different categories are shown in Fig.~\ref{fig:f3_2_heatmap}.

To achieve the joint cross similarity matrice from both same and different categories through addition or multiplication operations, we have adopted Euclidean distance instead of cosine similarity. We directly utilize the distance matrix D of different categories for coarse alignment with the similarity matrix S of the same category in terms of their numerical values, as shown in Eq.~\ref{eq:9}. By applying L2 normalization to the interaction matrix row-wise or column-wise, as in Eq.~\ref{eq:10}, we further achieve fine alignment.
\begin{small}
	\begin{equation}  
		\begin{split} 
			D_{cs\_ na}\left( {Q_{n},K_{a}} \right) = D_{cs\_ na}\left( {Q_{n},K_{a}} \right) \\
			- D_{cs\_ na}\left( {Q_{n},K_{a}} \right)_{min} \in \mathbb{R}^{l \times l} \label{eq:9}
		\end{split} 
	\end{equation}
	\begin{equation}  
		D_{cs\_ a}\left( {Q_{n},K_{a}} \right) = {\sum_{i = 1}^{l}\left( norm\left( {D_{cs\_ na}\left( {Q_{n},K_{a}} \right)}_{i,:} \right) \right)_{ij}} ~ \in \mathbb{R}^{1 \times l} \label{eq:10}
	\end{equation}
\end{small}

We found that adding matrices (Eq.~\ref{eq:11}) led to better performance than multiplying them. ${\gamma}$ is a hyperparameter and ${\theta}$ is a learnable parameter. The nonlinear activation function ${\tanh(\gamma\cdot\theta)}$ is used to limit the value to (-1,1), and more effective gradient propagation is obtained near the initial value ${\theta=0}$, enabling adaptive enhancement or suppression of matrix D on S. Finally, the vector ${{SD}_{cs\_ a}}$ is used to replace ${{S}_{cs\_ a}}$ in Eq.~\ref{eq:6}.
\begin{small}
	\begin{equation} 
		\begin{split}  
			{SD}_{cs\_ a}\left( {Q_{p},K_{a},Q_{n}} \right) = S_{cs\_ a}\left( {Q_{p},K_{a}} \right)\\
			+ \left( {1 + {\tanh(\gamma\cdot\theta)}} \right) \cdot D_{cs\_ a}\left( {Q_{n},K_{a}} \right) \in \mathbb{R}^{1 \times l} \label{eq:11}
		\end{split}	
	\end{equation}	
\end{small}

The Triplet loss\cite{Florian2015FaceNet} relies on comparing distances between embedding representations. Our method utilizes CSA in conjunction with the supervised classifier ${Cls_{cross}}$ to achieve two objectives at the fine-grained feature level: it brings the distances between images of the same class closer (enhancing fine-grained features with high similarity) and pushes the distances between images of different classes farther away (weakening fine-grained features with high similarity, i.e., small distances). 

The backbone of the anchor and pos branches are directly connected to the ${Cls_{base}}$ for supervised learning. If the classifier is not added to the neg branch, gradient conflicts will arise among the anchor, pos, and neg branches. Considering that the cross similarity modules S and D learn different content, the neg branch cannot be frozen during training (See the Appendix). Through various experimental attempts, it has been difficult to train effective models using networks constructed with triplet. 

To avoid gradient conflicts, the pos, anchor, and neg branches need to have equal status throughout the network, with each branch connected to the supervised classifier ${Cls_{base}}$. Ultimately, we ingeniously added another negative image, forming the four branches into a centrally symmetric closed-loop network called Quadruplet Cross Similarity(QCS) Network, thereby avoiding gradient conflicts and enhancing training stability, as shown in Fig.~\ref{fig:framework}. 

Meanwhile, the QCS offers computational efficiency. It allows us to calculate only two S matrices for the same category and two D matrices for different categories, with each interaction matrix utilized by two adjacent images simultaneously. The features obtained from the backbone without the class token go through an FC layer to produce vectors ${QK}$ and ${V}$. When calculating the interaction matrices with the features of two adjacent images, the weights of ${QK}$ are shared and act as either query ${Q}$ or key ${K}$. Consequently, the interaction matrices exhibit symmetry, where ${(Q2K1)^T=Q1K2}$ and ${(Q3K4)^T=Q4K3}$. When ${SD}$ fuses the matrices ${S}$ and ${D}$ by rows or columns, it reduces the need for two interactive calculations.

\subsection{Contrastive Residual Distillation}
The training loss curve in Fig.~\ref{fig:f4_3_res} shows a significant performance gap between ${{Cls}_{base}}$ and the enhanced ${{Cls}_{cross}}$ by the cross module. During inference, only one ${{Cls}_{base}}$ (red branch in Fig.~\ref{fig:framework}) is retained to predict results, posing a challenge in effectively transferring cross module information into the inference network.

A straightforward approach is to introduce residual connections between ${{Cls}_{base}}$ and ${{Cls}_{cross}}$ for direct gradient feedback. We explored diverse connection methods, such as element-wise addition followed by Global Average Pooling (GAP) and Bilinear pooling (BP) to fuse and pool the feature vectors. Additionally, we considered treating ${{Cls}_{cross}}$ as the teacher model and ${{Cls}_{base}}$ as the student model, introducing a Kullback-Leibler (KL) divergence loss or L2 loss to construct a self-distillation learning\cite{Linfeng2019BeYourOwnTeacher, Jie2020Residual}. We further integrated a ViT module into ${{Cls}_{cross}}$ to align with ${{Cls}_{base}}$, minimizing gradient conflicts on feature maps. We term the transfer of information from interactive contrastive learning in the cross module to the base network as Contrastive Residual Distillation.

\begin{small}
	\begin{equation}  
		\mathcal{L}_{total} = {\sum_{i = 1}^{C}\mathcal{L}_{{ce}_{base\_ i}}} + \lambda_{1} \cdot \mathcal{L}_{{ce}_{cross\_ i}} + \left\lbrack(\lambda_{2} \cdot \mathcal{L}_{{kl}_{i}}) or (\lambda_{3} \cdot \mathcal{L}_{{2}_{i}})\right\rbrack \label{eq:3_3_2}
	\end{equation}	
\end{small}

The joint training loss, utilizing cross entropy for all classifiers, is defined in Eq.~\ref{eq:3_3_2}. C is set to 2 for DCS and 4 for QCS, \(\lambda_{1}\) is set to 1, The [] indicates an optional element.


\section{Experiments}
\subsection{Datasets}
\textbf{RAF-DB}\cite{shanli2017reliable} is a real-world affective face database which contains 29,672 facial images labeled with basic or compound facial expressions. In the experiment, 15,331 images with 7 basic expressions(i.e. surprise, fear, disgust, happiness, sadness, anger, neutral) are choosed , of which 12271 are used for training and 3068 for testing.

\textbf{FERPlus}\cite{Ferlabel2016} is extended from FER2013(consists of 28,709 training images and 3,589 test images) and relabeled with 8 expression labels (neutral, happiness, surprise, sadness, anger, disgust, fear, contempt) by 10 taggers. In the experiment, we follow the code\cite{Ferlabel2016} to handle majority voting, which keeps a single label for each image and ignores the samples with unknown or NF labels, then measure prediction accuracy on the test set.

\textbf{AffectNet}\cite{Ali2017Affectnet} is a large-scale FER dataset in the wild, which contains about 440,000 manually annotated facial images. We utilized AffectNet-7/8 with 7/8 expressions, comprising 283,901/287,651 training and 3,500/4,000 validation images. Balance-sampling\cite{Imbalanced2019} is used since the training set exhibits a long-tailed distribution, while the validation set is balanced.

\subsection{Implementation Details}
We adopted POSTER++ removing the landmark branch as the baseline and constructed our model upon it. All face images are resized to 224${\times}$224 pixels for training and testing. We use IR50\cite{Jiankang2019Arcface}, which is pretrained on the Ms-Celeb-1M dataset\cite{Yandong2016Msceleb1m}, as the backbone. We employ an Adam optimizer with Sharpness Awareness Minimization \cite{Pierre2020Sharpness} to train the model (200 epochs for RAF-DB, 150 epochs for FERPlus and 80 epochs for AffectNet). For the DCS, the batch size is set to 48, the learning rate is initialized as 9e-6 and an exponential decay learning rate scheduling with a gamma of 0.98 is employed. For the QCS, the batch size and the initial learning rate are reduced to around half. It is trained end-to-end on a single Nvidia RTX3090 GPU via Pytorch. Each experiment is repeated five times, then the maximum accuracy is reported.

\begin{figure*}[t]
	\centering
	\begin{subfigure}[t]{0.22\textwidth}
		\includegraphics[width=\textwidth]{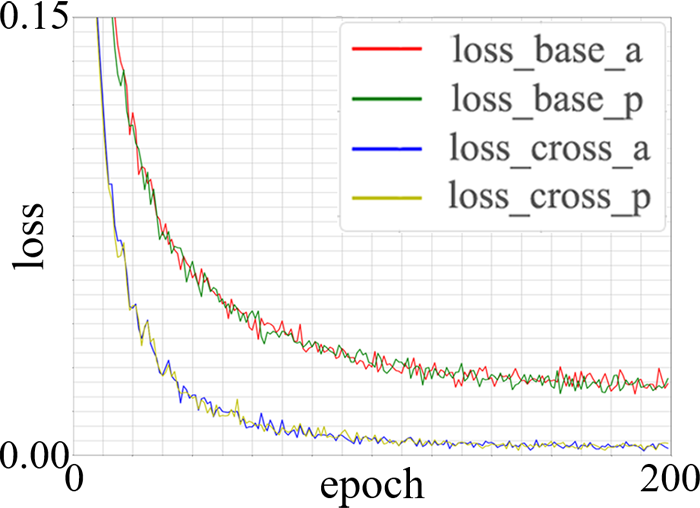}
		\caption{row 6}
		\label{fig:f4_3_res_c}
	\end{subfigure}    
	\hspace{1mm}
	\begin{subfigure}[t]{0.22\textwidth}
		\includegraphics[width=\textwidth]{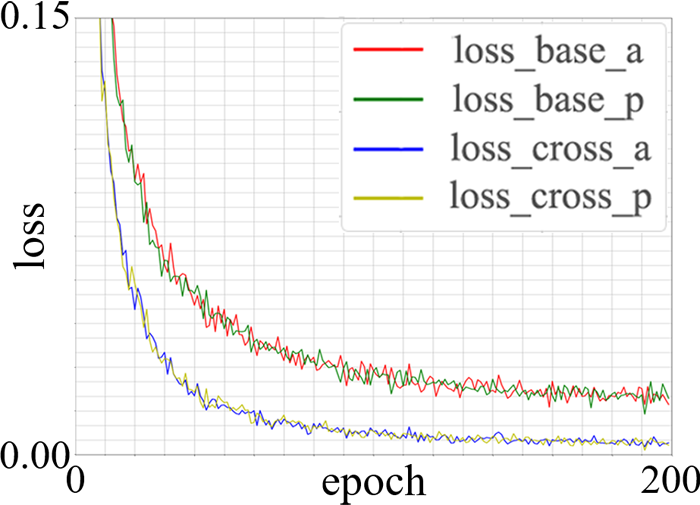}
		\caption{row 7}
		\label{fig:f4_3_res_d}
	\end{subfigure}  
	\hspace{1mm}
	\begin{subfigure}[t]{0.22\textwidth}
		\includegraphics[width=\textwidth]{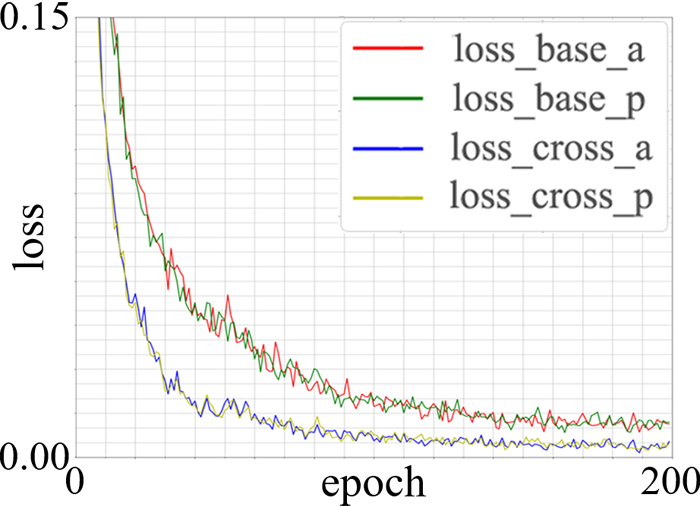}
		\caption{MSE}
		\label{fig:f4_3_sd_mse}
	\end{subfigure}    
	\hspace{1mm}
	\begin{subfigure}[t]{0.22\textwidth}
		\includegraphics[width=\textwidth]{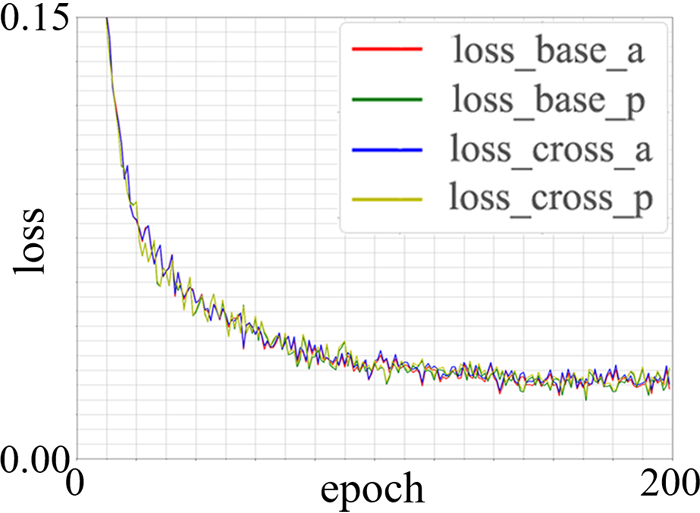}
		\caption{D matrix}
		\label{fig:f4_3_sd_d}
	\end{subfigure}    
  
	\caption{The training loss curves of ${Cls_{base}}$ and ${Cls_{cross}}$. (a)(b) correspond to the models in rows 6-7 of Table~\ref{tab:4_3_1}, respectively. (c)(d) correspond to the QCS, displaying only the anchor and pos branches.}
	\label{fig:f4_3_res}
\end{figure*}

\subsection{Ablation Analysis}
Initially, we trained a classification network utilizing the IR50+ViT backbone as a baseline. Row 1 of Table~\ref{tab:4_3_1} presents the accuracy of the baseline. Then, the DCS Network with shared weight branches was constructed on the backbone to evaluate the performance impact of each component in our method through ablation studies on RAF-DB and FERPlus.

\begin{table}[t]    
	\centering  
	\fontsize{9pt}{10pt}\selectfont
	\renewcommand{\arraystretch}{1.15}  
	\setlength{\tabcolsep}{1.8pt}   
	\begin{tabular}{c|c|c c|c c c|c|c c}   
		\hline    
		& IR50 & \multirow{2}{*}{SDPA} & \multirow{2}{*}{CSA} & ${\oplus}$ & \multirow{2}{*}{BP} & ${\oplus}$  & \multirow{2}{*}{KL} & \multirow{2}{*}{RAF-DB} & \multirow{2}{*}{FERPlus} \\     
		& ViT & & & GAP & & ViT & & & \\    
		\hline    
		1 & \checkmark & & & & & & &  92.05 & 91.06 \\    
		2 & \checkmark & & \checkmark & & & & & 91.98 & 90.83 \\
		3 & \checkmark & \checkmark & & \checkmark & & & & 91.95 & 90.93 \\
		4 & \checkmark & & \checkmark & \checkmark & & & & 92.14 & 91.02 \\
		5 & \checkmark & & \checkmark & & \checkmark & & & 92.24 & 91.25 \\
		6 & \checkmark & & \checkmark & & & \checkmark & & 92.57 & 91.22 \\
		7 & \checkmark & & \checkmark & & \checkmark & & \checkmark & 92.21 & 90.93 \\
		\hline    
	\end{tabular}%
	\caption{Accuracy (\%) comparison of various components on DCS. SDPA denotes Scaled Dot-Product Attention in ViT, GAP denotes  global average pooling, BP denotes bilinear pooling, KL denotes  KL divergence loss.  ${\oplus}$ denotes element-wise addition.}  
	\label{tab:4_3_1}%
\end{table}%

\paragraph{Evaluation of cross attetions.}
Rows 3 and 4 show that our CSA outperforms SDPA in the comparison. For a fair comparison, all structures remain consistent except for varying core interaction approaches, with Multi-Head Attention and class token removed for maximum structural simplicity. Features (7${\times}$7${\times}$c) extracted from 3 abstraction levels of the backbone interact individually and are concatenated on the spatial dimension. CSA avoids computing similarities between features of different levels, and SDPA adopts the same approach. The FC and MLP (Multi-Layer Perceptron) of the interaction module share weights across the interaction branches. The visualization results are shown in the Row 2 and Row 3 of Fig.~\ref{fig:visual_atten1}.

\paragraph{Evaluation of residual connection.}
We observe the performance gap between ${Cls_{base}}$ and ${Cls_{cross}}$ through their training loss curves (Fig.~\ref{fig:f4_3_res}), yet we are unable to obtain the accuracy of ${Cls_{cross}}$ during the inference stage. The model accuracy in row 2 of Table~\ref{tab:4_3_1} is lower than the baseline(row 1), indicating the gradient differences after multiple layers may have distorted the baseline model's original feature representation. The improvement from rows 2 to 4 of Table~\ref{tab:4_3_1} demonstrates the information transfer through residual connections. Rows 4-6 suggest superior performance from minimizing gradient conflicts in connections, that BP retains fine-grained features while ViT module align ${Cls_{cross}}$ with ${Cls_{base}}$. The experiments below adopt the ViT connections.

\paragraph{Evaluation of cross similarity module overfitting.}
In Eq.~\ref{eq:3_3_2}, one of \(\lambda_{2}\) and \(\lambda_{3}\) is set to 1, while the other is set to 0. As shown in Table~\ref{tab:4_3_1}(row 7) and Fig.~\ref{fig:f4_3_res_d}, adding the KL divergence loss narrows the losses gap between ${Cls_{base}}$ and ${Cls_{cross}}$, but the accuracy of the model decreases. This phenomenon becomes even more evident in the more complex QCS network using MSE loss, as shown in Fig.~\ref{fig:f4_3_sd_mse}. It is possible that the cross module exhibits overfitting, influencing the base model's generalization capacity. We set the dropout ratio ranging from 0.1 to 0.4 in the ViT modules to mitigate overfitting during training on AffectNet.

\paragraph{Evaluation of intra-class similarity and inter-class dissimilarity.}
To explore how intra-class similarity and inter-class differences influence effective feature extraction, we conducted experimental analysis leveraging the QCS Network. In Table~\ref{tab:4_3_3}, S represents retaining only the intra-class similarity matrix and D represents retaining only the inter-class distance matrix. SD utilizes Eq.~\ref{eq:11} to fuse the interaction matrices S and D, \textgamma=1.

\begin{table}[t]  
	\centering  
	\fontsize{9pt}{10pt}\selectfont
	\renewcommand{\arraystretch}{1.1}
	\begin{tabular}{c | c c c} 
		\hline  
		& S & D & SD \\  
		\hline  
		RAF-DB & 92.18 & 92.11 & 92.47 \\  
		FERPlus & 91.12  & 90.96  & 91.21  \\  
		\hline  
	\end{tabular}  
	\caption{Comparison(\%) of different cross matrices.}  
	\label{tab:4_3_3}  
\end{table}
Table~\ref{tab:4_3_3} shows S contributes more to improving the model performance compared to D. As can be seen from Fig.~\ref{fig:f4_3_sd_d}, ${Cls_{cross}}$, which is enhanced by D, does not show loss reduction compared to ${Cls_{base}}$. Fig.~\ref{fig:f3_2_heatmap} shows that the inter-class similarity matrix has more dark blue areas (i.e., more bright yellow highlights in the inter-class distance matrix), indicating more dispersed attention. We infer that although weakening the features with small distances (high similarity) across categories helps separate redundant features, the simultaneously enhanced features with large distances do not necessarily contribute to classification. The SD matrix, resulting from the adaptive fusion of S and D, harnesses the strengths of both S and D to extract more clearly features.

\subsection{Comparison with the State of the Art}
Taking into account the potential overfitting issue in the cross module, we incorporated data augmentation such as random cropping, rotation and color jitter. We also optimized the sampling strategy for positive and negative samples, resulting in further improvements in model performance. (See the Appendix for details.)

Table~\ref{tab:sota} compares our best results with the current state-of-the-art methods on RAF-DB, FERPlus, and AffectNet-7/8. APViT and the other methods listed below are Transformer-based methods, while POSTER, POSTER++, and S2D incorporate additional landmark information. Our proposed method achieves these results solely by leveraging self-supervised contrastive learning to mine richer intrinsic information, without incorporating any extra landmark information.

While QCS effectively separates redundant features, its performance does not demonstrate a significant advantage compared to DCS. This may be due to the fact that more complex interactive modules may pose challenges during training, where over-sampling of underrepresented classes can exacerbate overfitting, ultimately resulting in a performance trade-off. Nevertheless, our method outperforms recent state-of-the-art methods on those datasets.

\begin{table}[t]
	\centering
	\fontsize{9pt}{10pt}\selectfont
	\renewcommand{\arraystretch}{1.1}
	\setlength{\tabcolsep}{2pt}
	\begin{tabular}{c | c c c c}
		\hline  
		Methods & RAF-DB & FERPlus & AffectNet-7 & AffectNet-8\\
		\hline  
		SCN\shortcite{Kai2020Suppressing}  & 87.03 & 88.01 & - & 60.23\\
		RUL\shortcite{Yuhang2021Relative}  & 88.98 & - & - & -\\
		DMUE\shortcite{Jiahui2021DiveintoAmbiguity}  & 89.42 & 89.51 & - & 63.11\\
		TransFER\shortcite{fanglei2021transfer}  & 90.91 & 90.83 & 66.23 & -\\
		Face2Exp\shortcite{Dan2022Face2Exp}  & 88.54 & - & 64.23 & -\\
		EAC\shortcite{YuhangZhang2022EAC}  & 89.99 & 89.64 & 65.32 & -\\
		APViT\shortcite{fanglei2022vision}  & 91.98 & 90.86 & 66.91 & -\\
		TAN\shortcite{Fuyan2023Facial}  & 90.87 & 91.00 & 66.45 & -\\ 
		POSTER\shortcite{Zheng2023Poster}  & 92.05 & \underline{91.62} & 67.31 & 63.34\\
		POSTER++\shortcite{Jiawei2023POSTERV2}  & 92.21 & - & 67.49 & 63.77\\
		S2D\textdaggerdbl\shortcite{Yin2023From}  & \underline{92.57} & 91.17 & 67.62 & 63.06\\
		\hline		
		DCS(Ours) & \underline{92.57} & 91.41 & \underline{67.66} & \textbf{64.40}\\
		QCS(Ours) & 92.50 & 91.41 & \textbf{67.94} & \underline{64.30}\\
		QCS(Ours)\textdaggerdbl & \textbf{93.02} & \textbf{91.85} & - & -\\
		\hline  
	\end{tabular}%
	\caption{Performance comparison (\%) on RAF-DB, FERPlus, and AffectNet. \textdaggerdbl means pre-trained on AffectNet-8.}
	\label{tab:sota}%
\end{table}%

\subsection{Visualization}
\begin{figure}[t]  
	\centering  
	\includegraphics[width=8.5cm]{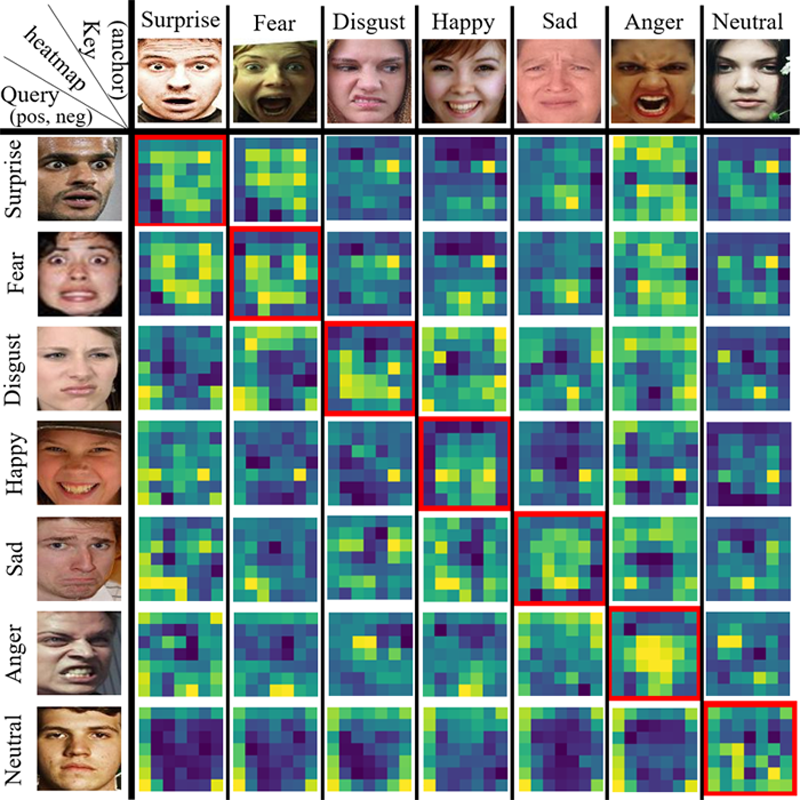}  
	\caption{The confusion matrix of Cross Similarity heatmaps on RAF-DB. The anchor image located above a certain column is used as the key to calculate cross similarity on that anchor. The yellow highlighted area represents high similarity. During adaptive fusion through Eq.~\ref{eq:11}, high-similarity features on the confusion matrix diagonal are enhanced, while those outside the diagonal are suppressed.}  
	\label{fig:visualization_1}  
\end{figure}

\paragraph{Cross Similarity Visualization}
To visually demonstrate the effectiveness of QCS in extracting distinctive features and separating redundant features, we calculated the cross similarity between every pair of the 7 expressions. We computed the similarity matrices between the anchor and positive as well as between the anchor and negative.  Fig.~\ref{fig:visualization_1} only displays the similarity heatmaps of the anchor that is used as Key, and the heatmaps for positive and negative are not shown.

The heatmaps in diagonal positions show that the distinctive features extracted from images of the same category mainly concentrate on the areas of eyes and mouth corners. The heatmaps in non-diagonal positions show the redundant features separated from comparing the images in a certain column (Key) with various expressions. As can be seen from the last row of heatmaps, Neutral, which does not have the facial features of expressions, mainly separates out feature regions from various expressions in the corners of the images.

\begin{figure}[t]  
	\centering  
	\includegraphics[width=3.2in]{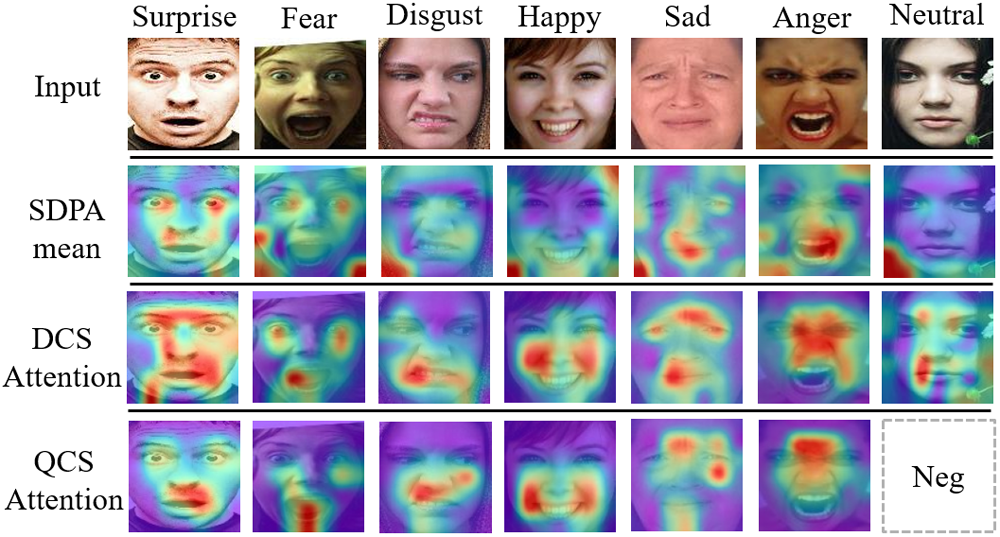}  
	\caption{Cross Attentions visualization on RAF-DB. The QCS uniformly employs the Neutral image as the input of negative branch.}  
	\label{fig:visual_atten1}  
\end{figure}

\paragraph{Attention Visualization}
We conducted a visual comparison of the cross attentions in the Ablation Analysis. Due to the removal of the class token for cross module, we present SDPA by averaging the attention weights from all other tokens.From Row 2 of Fig.~\ref{fig:visual_atten1}, it can be observed that the attention areas of SDPA are not prominent, and a significant number of redundant features in the corner regions of the images receive greater emphasis during information transmission. In contrast, CSA directly acts between the input and output feature vector of its own branch, enabling more direct feedback. As clearly shown in Row 3 of Fig.~\ref{fig:visual_atten1}, our method achieves effective attention capture, with the effects mainly concentrated on key areas such as the facial features, cheeks, and eyebrows.

Rows 3 and 4 of Fig.~\ref{fig:visual_atten1} demonstrate the attentions of the cross module S in the DCS Network and the cross module SD in the QCS Network, respectively. It can be observed that the highlighted areas in QCS Attention are more concentrated, the local areas of the eyes are weakened, more attention is being paid to the movement of facial muscles. For instance, col. 4 (Happy) emphasizes the cheek area, while col. 2 (Fear), col. 3 (Disgust), and col. 5 (Sad) focus on the cheekbone area. Additionally, both col. 6 (Anger) and col. 5 (Sad) highlight the point between the eyebrows.

\section{Conclusion}

This paper proposes a feature refining network named Quadruplet Cross Similarity (QCS) to mine richer intrinsic information for Facial Expression Recognition (FER). The QCS comprises a closed-loop symmetric network for stable joint training. It adaptively extracts discriminative features within the same class and synchronously separates interfering features between different classes. The cross module is based on Cross Similarity Attention (CSA), a novel global spatial attention designed for different image pairs, which directly assigns similarity attention between the input and output feature vectors of a branch. Extensive experiments on several public datasets indicate that our QCS effectively refines the features into a purer form and achieves state-of-the-art performance. Furthermore, effectively transferring the performance of the cross module back to the basic inference network while avoiding overfitting, and enhancing the stability of joint training, remains a crucial task.

\section{Acknowledgments}
This work was supported by funding from the National Natural Science Foundation of China (U21B2035, 82372218).

\newpage

\bibliography{aaai25}

\newpage

\section{Appendix}

\begin{figure}[h]  
	\centering  
	\includegraphics[width=3.2in]{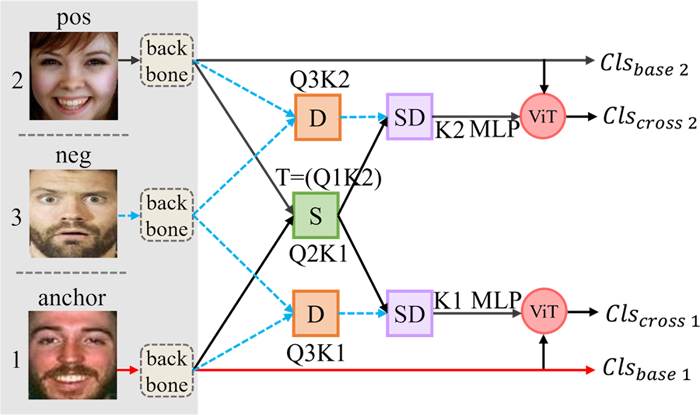}  
	\caption{Challenges in constructing a stable training Triplet Network.}  
	\label{fig:triplet_network}  
\end{figure}

\subsection{Failed Attempt for the Triplet Network}

Unlike Triplet loss \cite{Florian2015FaceNet}, our network employs CSA in conjunction with a supervised classifier to achieve the purpose of bringing similar classes closer and pushing different classes farther away at the fine-grained feature level. If the neg branch is not connected to the same classifiers ${Cls_{base}}$ and ${Cls_{cross}}$ as the anchor and pos branches, gradient conflicts will arise among different branches during training. Due to the different inputs for the cross module S and the cross module D, they are unable to share weights. If the neg branch is frozen, as illustrated by the blue dashed arrow in Fig.~\ref{fig:triplet_network}, then the cross module D cannot be effectively learned. Despite conducting numerous experiments, we were unable to train an effective model.

\subsection{Concat and Split on Spatial Dimension}

\begin{figure}[t]  
	\centering  
	\begin{subfigure}[t]{0.4\textwidth} %
		\includegraphics[width=\textwidth]{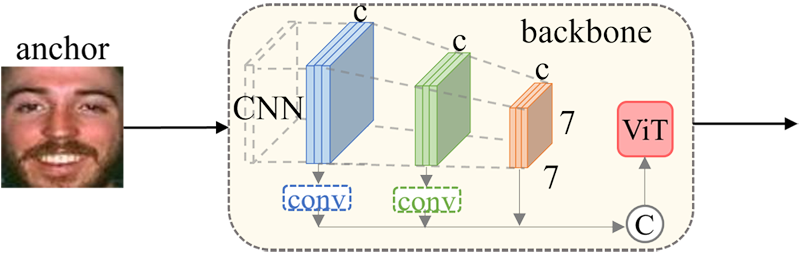}  
		\caption{}  
		\label{fig:backbone}  
	\end{subfigure}    
	\begin{subfigure}[t]{0.38\textwidth}  
		\includegraphics[width=\textwidth]{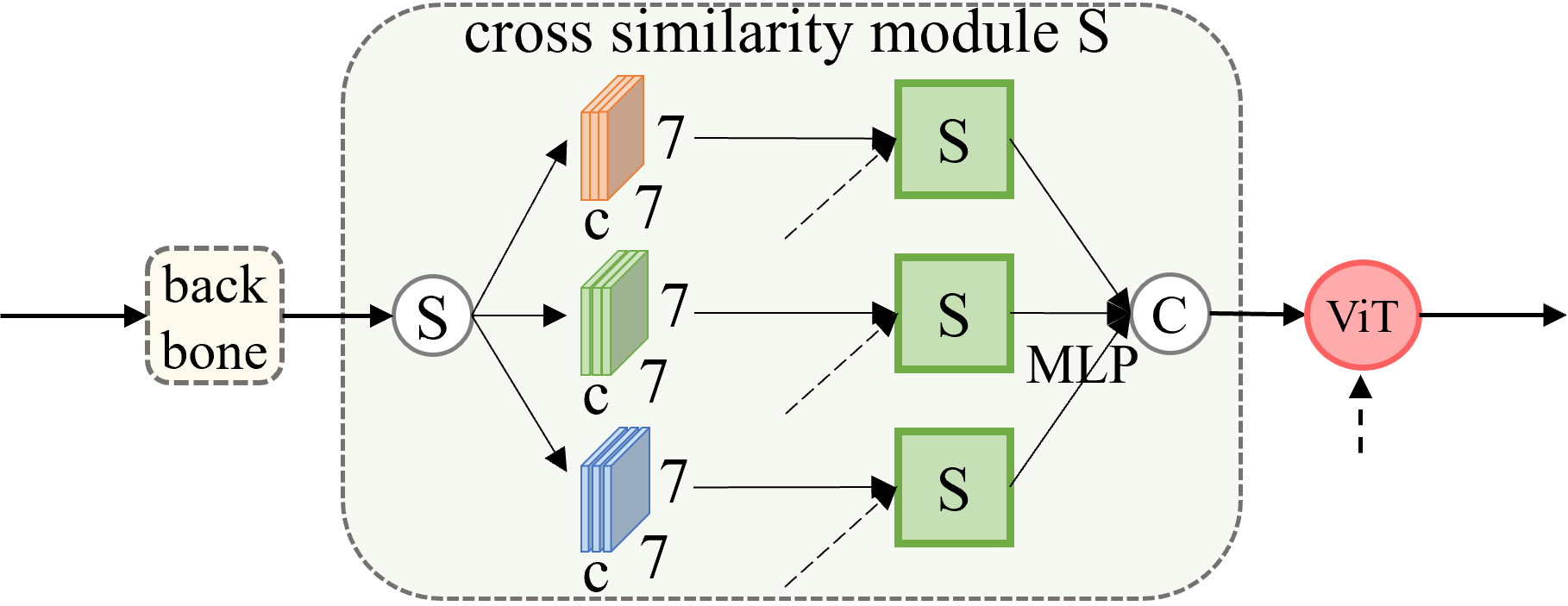}  
		\caption{}  
		\label{fig:concat_split}  
	\end{subfigure}  
	\caption{The concat and split operations on the backbone and cross similarity module.  \textcircled{c} denotes concat on spatial dimension, \(\circledS\) denotes split on spatial dimension. The dashed arrow represents the omitted network.}
	\label{fig:backbone_concat}  
\end{figure}

In the backbone (See Fig.~\ref{fig:backbone}), features (7${\times}$7${\times}$c) extracted from three abstraction levels are concatenated along the spatial dimension before being fed into the standard ViT for computing interactions. Our CSA avoids computing similarities between features of different levels. In the cross similarity module (See Fig.~\ref{fig:concat_split}), features (7${\times}$7${\times}$c) extracted from three abstraction levels of the backbone interact individually and are then concatenated along the spatial dimension. In the experimental section, we only visualize the cross similarity and attention from the highest abstraction level.

\subsection{Training Improvement and Results Details}
\begin{figure*}[t]
	\centering
	\begin{subfigure}[b]{1.0\textwidth}
		\centering
		\includegraphics[width=0.245\textwidth]{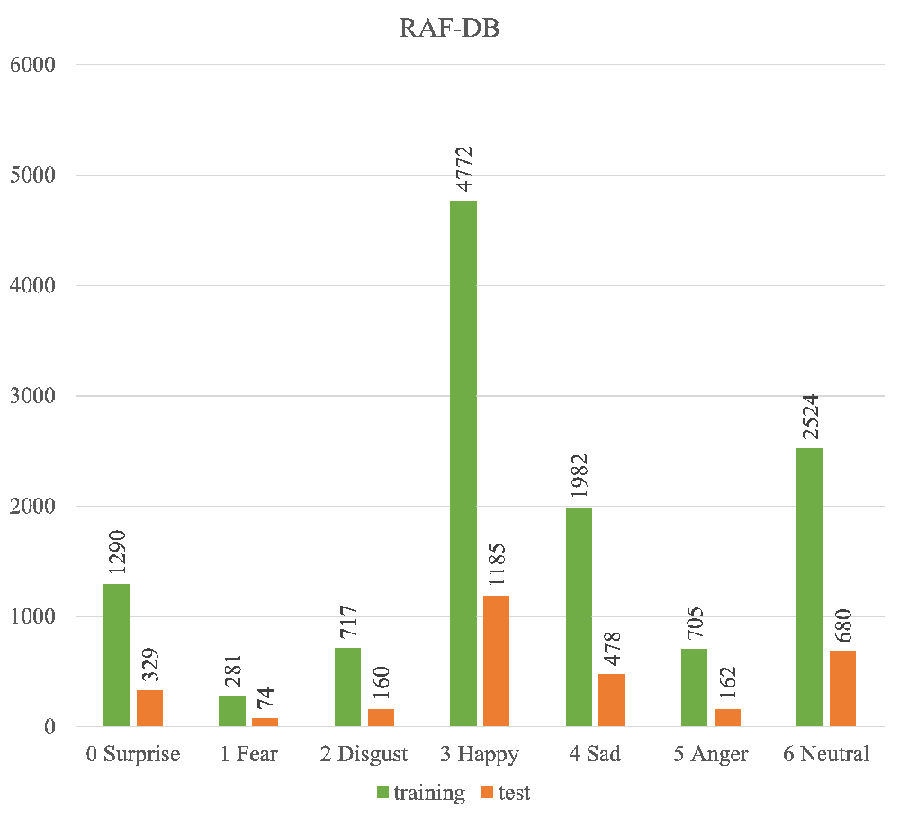}
		\hfill
		\includegraphics[width=0.245\textwidth]{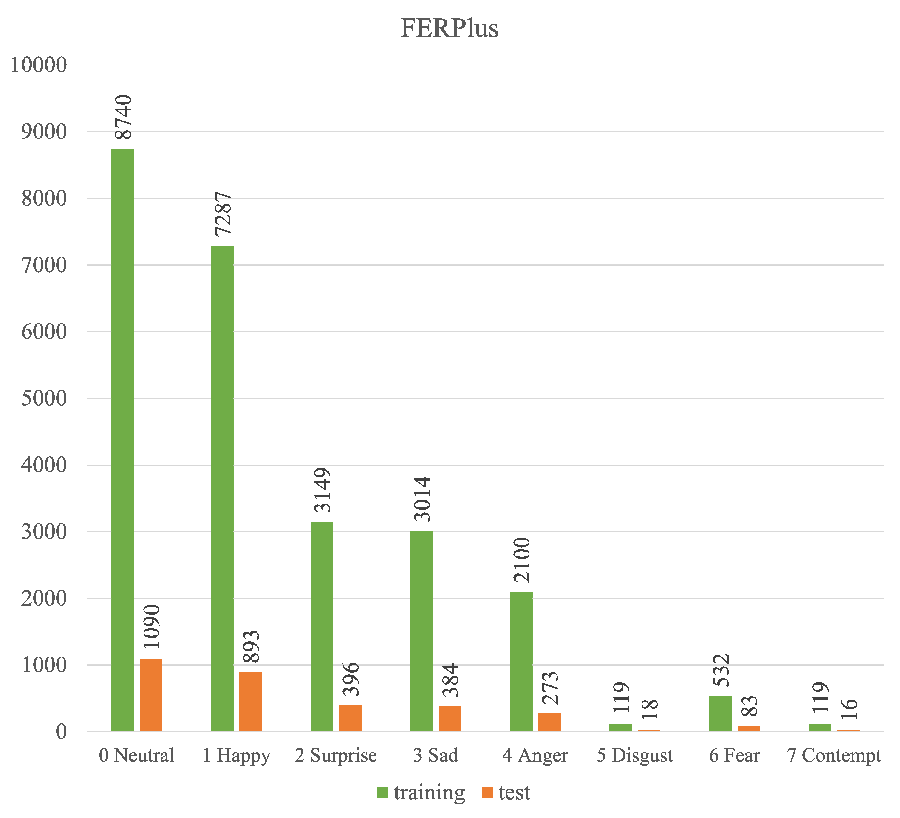}
		\hfill
		\includegraphics[width=0.245\textwidth]{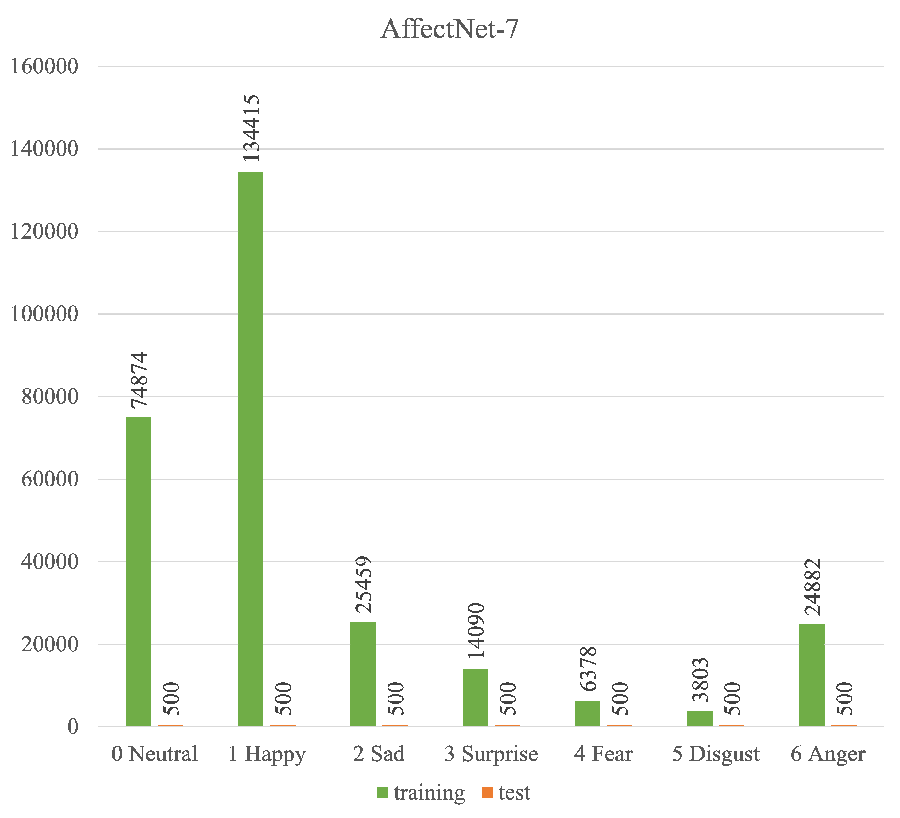}
		\hfill
		\includegraphics[width=0.245\textwidth]{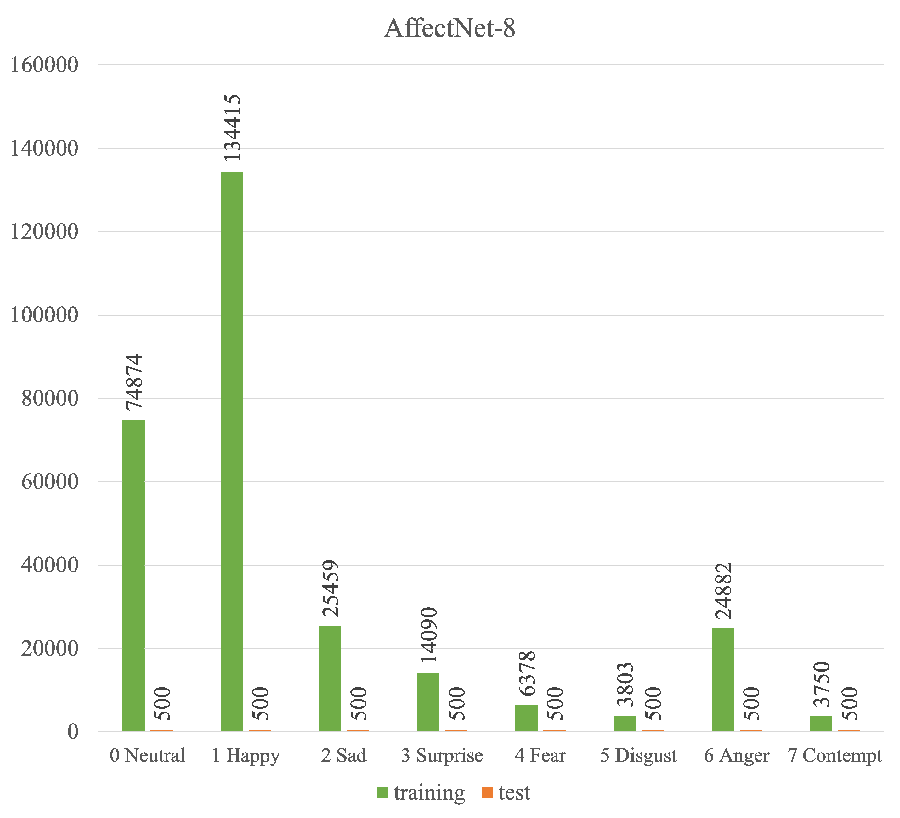}
		\caption{sample distributions}
		\label{fig:sample_distribution}
	\end{subfigure}
	\vskip\baselineskip
	\begin{subfigure}[b]{1.0\textwidth}
		\centering
		\includegraphics[width=0.245\textwidth]{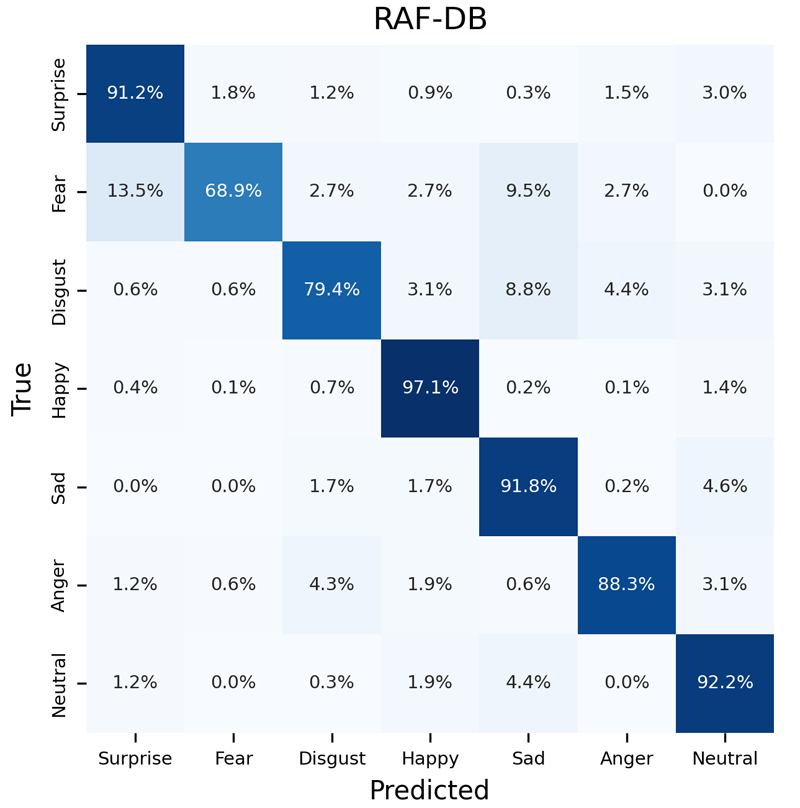}
		\hfill
		\includegraphics[width=0.245\textwidth]{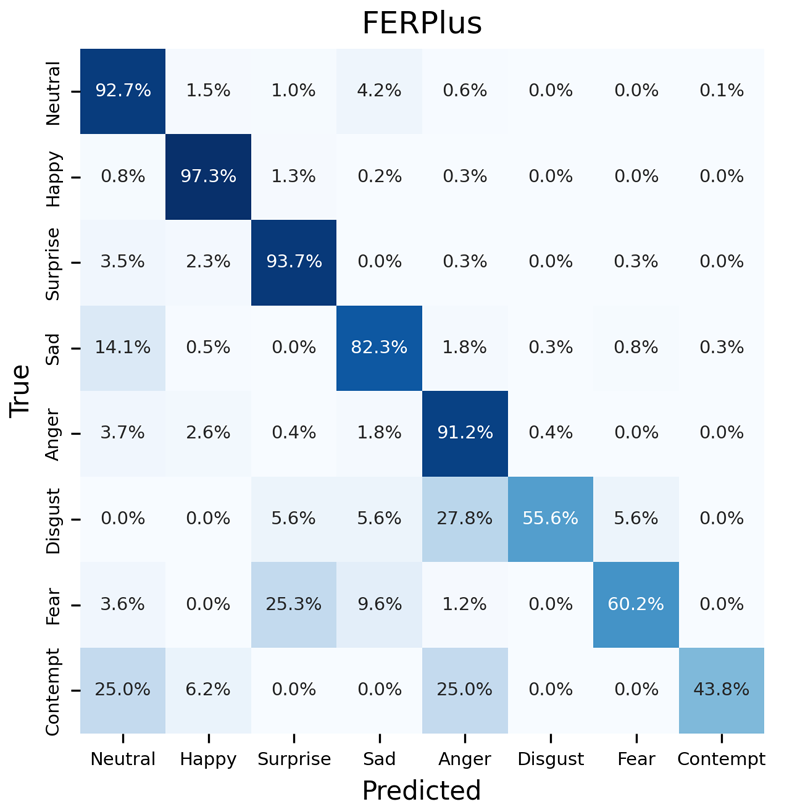}
		\hfill
		\includegraphics[width=0.245\textwidth]{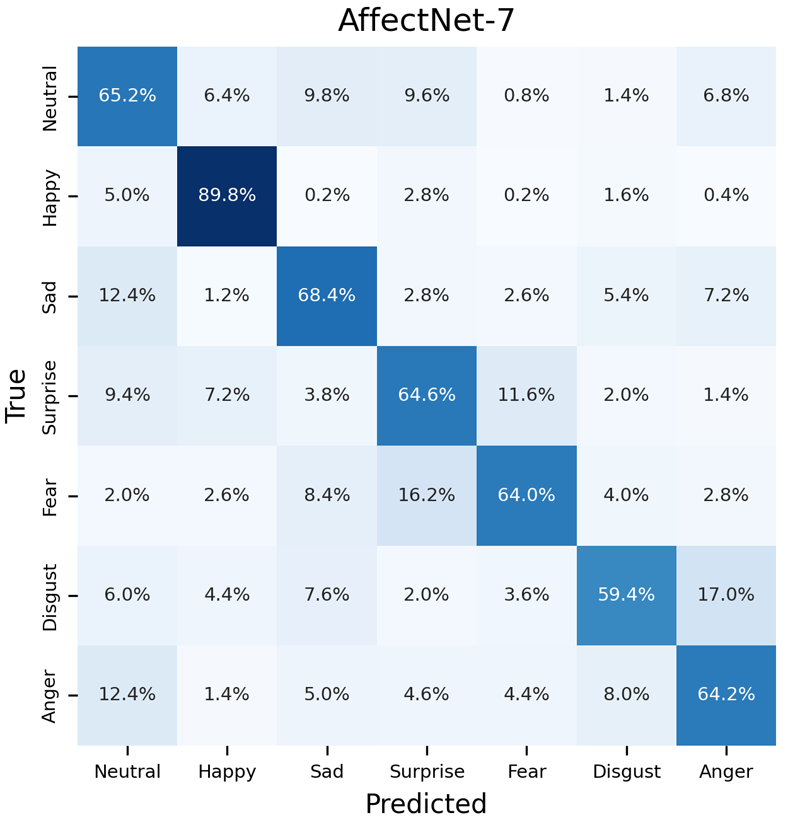}
		\hfill
		\includegraphics[width=0.245\textwidth]{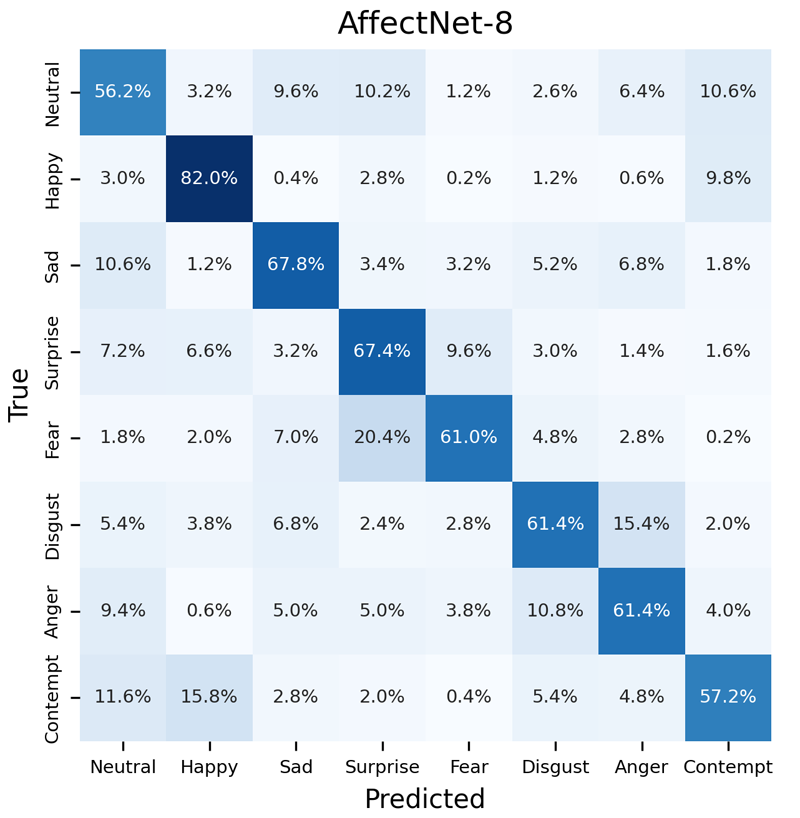}
		\caption{confusion matrices}
		\label{fig:confusion_matrix}
	\end{subfigure}
	\caption{Details of sample distribution and confusion matrices of QCS on each dataset.}
	\label{fig:dataset_matrix}
\end{figure*}

As can be seen from Fig.~\ref{fig:sample_distribution}, the distribution of samples across different categories within each dataset is highly unbalanced, and there are also significant variations in the proportion of training and testing samples among different datasets. Taking into account the potential overfitting issue in the cross module of the framework, we adopted different parameter settings and sampling strategies when training on different datasets. Fig.~\ref{fig:confusion_matrix} displays the accuracy confusion matrices for each test set, where the accuracy of each category correlates with the number of samples in the training set.

\subsubsection{RAF-DB}
The ratio of training to testing images for each category is approximately 4:1. Since the test set and the training set exhibit similar unbalanced distributions, the relatively lower accuracy rates for the categories of Disgust and Fear do not significantly affect the overall accuracy. For positive samples, random sampling is employed. After determining the positive samples in QCS, negative samples are randomly sampled from all samples of the other categories. The learning rate is initialized as 9e-6 for DCS and 4e-6 for QCS. When fine-tuning QCS on the pre-trained model QCS-Affect-8, the learning rate is initialized to 3e-6.

\subsubsection{FERPlus}
The ratio of training to testing images for each category is approximately 8:1. Although, like the RAF-DB dataset, the test set and the training set exhibit similar unbalanced distributions, the extreme scarcity of samples in the Disgust and Contempt categories may exacerbate the overfitting issue specifically on these categories. The test results from the confusion matrix also indicate that the model's accuracy is very low for the Disgust, Fear, and Contempt categories. 

For positive samples, random sampling is employed. After determining the positive samples in QCS, negative samples are randomly sampled from all samples of the other categories. To mitigate overfitting, we intensified our sample augmentation efforts by performing random crops of 224x224 within images of 234x234, applying random rotations of up to 10 degrees, and randomly jittering brightness, contrast, saturation, and hue. Additionally, we set the dropout rate of ${{ViT}_{cross}}$ to 0.2 to stabilize the training results for DCS and QCS. The learning rate is initialized as 8e-6 and 3.5e-6 for QCS. When fine-tuning QCS on the pre-trained model QCS-Affect-8, the learning rate is initialized to 2.5e-6.

\begin{table}[t]
	\centering
	\fontsize{9pt}{10pt}\selectfont
	\renewcommand{\arraystretch}{1.1}
	\setlength{\tabcolsep}{2pt}
	\begin{tabular}{c | c | c c c}
		\hline  
		Methods & Backbone & RAF-DB & FERPlus & AffectNet-7\\
		\hline 
		POSTER(w/o lm) & IR50 & 90.51 & - & 64.95\\ 
		POSTER++(w/o lm) & IR50 & 91.62 & - & -\\
		S2D(w/o lm) & ViT-B/16 & 91.07 & 90.18 & 66.09\\
		\hline		
		QCS(Ours) & IR50 & 92.50 & 91.41 & 67.94\\
		\hline  
	\end{tabular}%
	\caption{Performance comparison (\%), (w/o lm) means without landmark.}
	\label{tab:landmark}%
\end{table}%

\subsubsection{AffectNet-7/8}
Given that the amount of training data has increased significantly compared to RAF-DB and FERPlus, we train model on the subset (only sample 1/N of the total samples) in one epoch, where N is 8 for DCS and 12 for QCS. The training will run for a total of 80 epochs.

Balance-sampling\cite{Imbalanced2019} is used since the training set is unbalanced, while the validation set is balanced. Positive samples are selected using Balance-sampling\cite{Imbalanced2019}. After determining the categories of positive sample in QCS, negative sample categories are randomly chosen with equal probability from the remaining categories, and then negative samples are randomly sampled within the selected category to achieve Balance-sampling for negative samples. 

Over-sampling minority classes in such a long-tailed distributed training set is prone to overfitting. To mitigate overfitting, we incorporate samples of the Fear, Disgust, and Contempt classes from RAF-DB and FERPlus for training. The dropout rate of ${{ViT}_{cross}}$ is set to 0.4 for DCS. We intensified our sample augmentation efforts by performing random crops of 224x224 within images of 236x236, applying random rotations of up to 12 degrees, and randomly jittering brightness, contrast, saturation, and hue. The learning rate is initialized as 9e-6 for DCS and 4e-6 for QCS.

\subsection{Comparison without landmark}

\begin{figure}[t]  
	\centering  
	\includegraphics[width=3.2in]{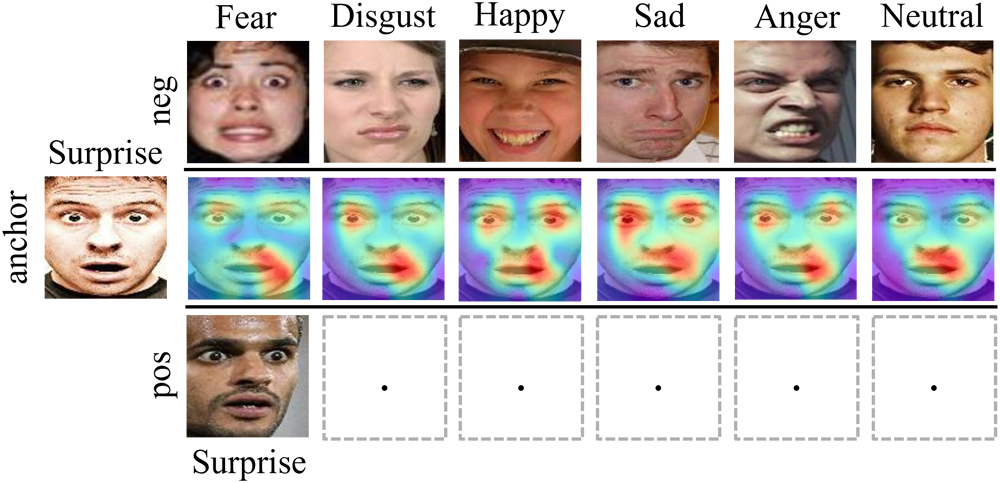}  
	\caption{QCS Attention visualization for anchor. The pos samples all use the same image, while the neg samples each use an image from different categories.}  
	\label{fig:visual_atten2}  
\end{figure}

Table~\ref{tab:landmark} compares the performance of our QCS to POSTER, POSTER++ and S2D, all without incorporating additional landmark information. Our QCS has demonstrated significant performance improvements. Compared with POSTER++, our method shows no increase in inference time when all interactive modules are removed and only a single branch is retained.

\subsection{Attention Visualization}

Fig.~\ref{fig:visual_atten2} displays the extraction of distinctive features from two images belonging to the same category (Surprise), while simultaneously isolating redundant features across samples from six other distinct categories to highlight the similarities and differences in attention. Almost all images exhibit enhanced areas between the nose and the corners of the mouth, likely due to significant variations in muscle movement within this region across different expressions. On the other hand, the eye region in column 1 (Fear) resembles that of Surprise and is notably weakened.

\end{document}